\documentclass[letterpaper, 10 pt, conference]{ieeeconf}  

\IEEEoverridecommandlockouts                              

\usepackage{graphicx} 
\usepackage{amsmath} 
\usepackage{amssymb}  
\usepackage[style=ieee,backend=bibtex]{biblatex}
\usepackage{booktabs}
\usepackage{esvect}
\usepackage{subcaption}
\bibliography{two_finger}  

\usepackage{todonotes}
\usepackage{xspace}
\usepackage{hyperref}
\usepackage{soul}
\usepackage{tikz}
\usetikzlibrary{arrows,automata}

\title{\LARGE \bf
Bio-Inspired Grasping Controller for Sensorized 2-DoF Grippers
}

\author{Luca Lach$^{1,2}$, Séverin Lemaignan$^{1}$, Francesco Ferro$^{1}$, Helge Ritter$^{2}$, Robert Haschke$^{2}$
\thanks{This work was supported by the European Union Horizon 2020 Marie Curie Actions under Grant 813713 NeuTouch.}
\thanks{$^{1}$PAL Robotics, Barcelona, Spain}%
\thanks{$^{2}$Neuroinformatics Group, Bielefeld University, Bielefeld, Germany}}%

\renewcommand{\vec}[1]{\mathbf{#1}} 
\newcommand{\T}{\top}

\newcommand{\ft}{\ensuremath{f_\theta}\xspace}
\newcommand{\fte}{\ensuremath{f_\phi}\xspace}
\newcommand{\g}{{\text{goal}}}
\newcommand{\inner}{{\text{int}}}
\newcommand{\ext}{{\text{ext}}}
\newcommand{\cmd}{{\text{cmd}}}
\newcommand{\ks}{\ensuremath{k}\xspace} 
\newcommand{\Dq}{\ensuremath{\Delta \vec q}\xspace} 
\newcommand{\Dx}{\ensuremath{\Delta x}\xspace} 
\newcommand{\oc}{{oc_i}}
\newcommand{\adj}{\ensuremath{Ad^\T_{T_\oc^{-1}}}\xspace}
\newcommand{\FC}{FC}

\newcommand{\fgoal}{\ensuremath{f^{\text{goal}}}\xspace} 
\newcommand{\fext}{\ensuremath{f^{\text{ext}}}\xspace} 
\newcommand{\fint}{\ensuremath{f^{\text{int}}}\xspace} 

\newcommand{\kp}{\ensuremath{K_{\text{P}}}\xspace}
\newcommand{\ki}{\ensuremath{K_{\text{I}}}\xspace}

\newcommand{\eef}{end-effector\xspace}
\newcommand{\eefs}{end-effectors\xspace}
\newcommand{\pjg}{parallel-jaw gripper\xspace}

\newcommand{\one}{A\xspace}
\newcommand{\two}{B\xspace}

\begin{document}

\maketitle
\thispagestyle{empty}
\pagestyle{empty}

\begin{abstract}
We present a holistic grasping controller, combining free-space position control and in-contact force-control for reliable grasping given uncertain object pose estimates. 
Employing tactile fingertip sensors, undesired object displacement during grasping is minimized by pausing the finger closing motion for individual joints on first contact until force-closure is established. 
While holding an object, the controller is compliant with external forces to avoid high internal object forces and prevent object damage.
Gravity as an external force is explicitly considered and compensated for, thus preventing gravity-induced object drift.
We evaluate the controller in two experiments on the TIAGo robot and its \pjg proving the effectiveness of the approach for robust grasping and minimizing object displacement.
In a series of ablation studies, we demonstrate the utility of the individual controller components.
\end{abstract}

\section{Introduction}
One of the most important skills service robots need to master is that of object manipulation.
Since the majority of object manipulation approaches are vision-based \cite{du2021vision}, they have a huge disadvantage compared to human grasping:
they lack environment feedback during physical object interaction and are therefore prone to high error rates.
This applies especially to mobile robots since they have to deal with more sources of uncertainty than fixed setups (e.g. unreliable hand-eye calibration).
Robotic \eefs equipped with tactile sensors can provide the robot with data about physical object interactions and thus enable more robust grasping approaches~\cite{Bartolozzi2016}.

By measuring and controlling grasp forces, manipulation tasks can be modelled and executed much more precisely.
In situations where fragile objects are being handled, the grasp force should be limited to prevent object damage and yet ensure a firm grasp.
Furthermore, undesired object movements during grasping should be minimized to avoid undesired side-effects such as tipping over the object.
One key concept in grasping that formalizes these characteristics is that of force closure.
It requires the robot to only apply forces on the object that result in a net-force of zero, hence avoiding object displacement.

Many force controllers have been proposed for different robot setups and arm-hand configurations that build upon force closure grasps.
However, most of them do not consider the effects of external forces on the controller behavior.
Mobile robots typically have to deliver objects to humans or place them in another location after grasping.
Thus, they also need to be able to correctly handle external forces in the form of gravity or humans interacting with the \eef.

In this work, we propose a novel grasping controller with the following characteristics:
\begin{enumerate}
    \item During the grasping process, the object is only minimally displaced, exploiting tactile feedback to account for uncertainties in visual object pose estimation.
    \item Holding the object, a desired grasp force is maintained, while also compliantly reacting to external forces.
    \item Gravity as an external force is estimated and explicitly considered to avoid undesired drift of the object.
    \item Small imbalanced force measurements are ignored, again avoiding undesired object drift.
\end{enumerate}

\begin{figure}
    \centering
    \includegraphics[width=\linewidth]{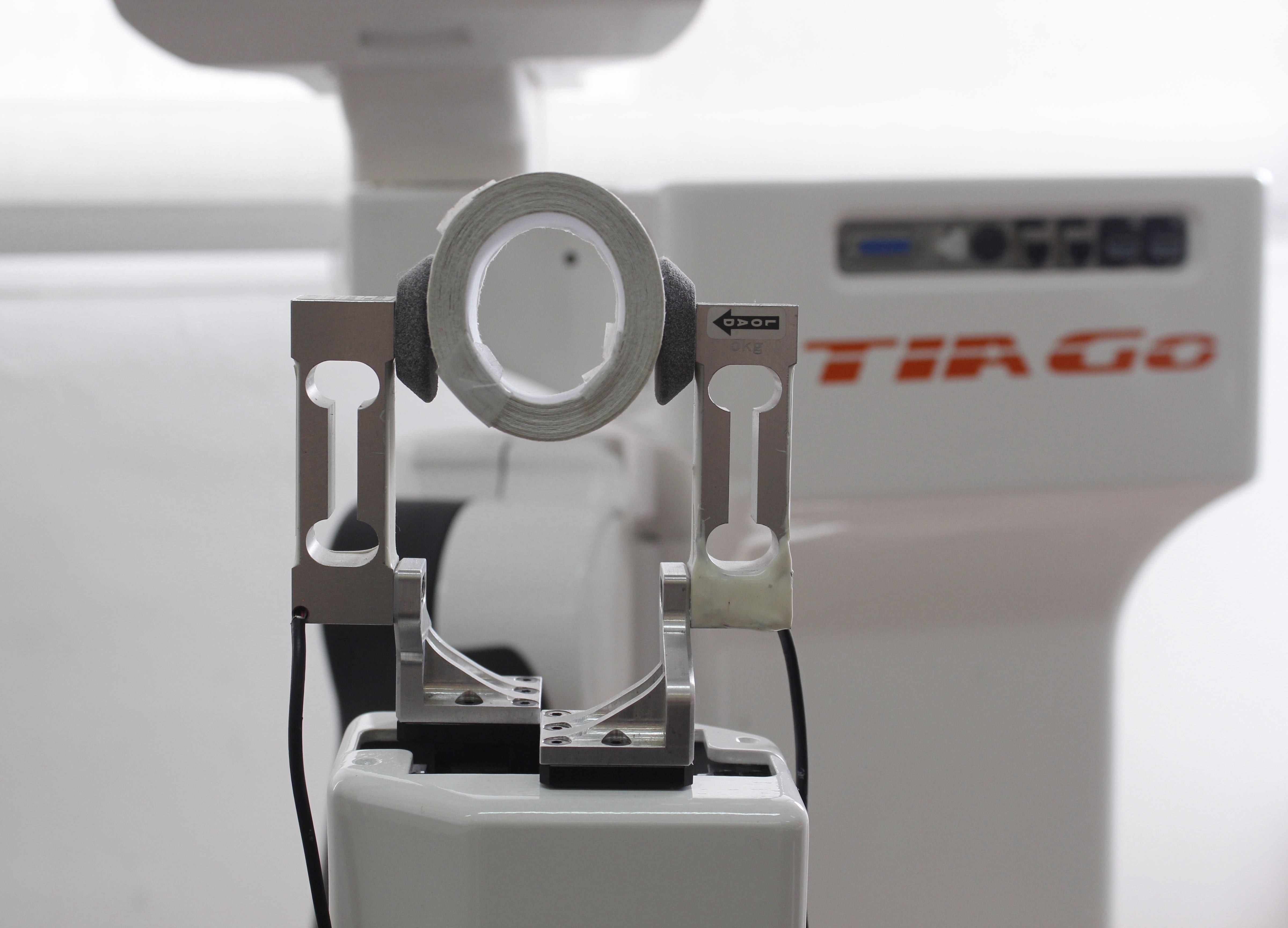}
    \caption{Grasping a soft object with TIAGo's sensorized \pjg using force control.}
    \label{fig:grasp}
\end{figure}

To the best of our knowledge, this is the first grasping controller, that explicitly compensates external forces while minimizing object displacement and accounting for gravity.
We first derive the core concepts of our controller from human grasping strategies and formulate them in a general fashion. Subsequently we derive an instance specifically tailored for 2-DoF grippers.
Finally, we validate them on a sensorized \pjg of the TIAGo service robot as shown in Fig.~\ref{fig:grasp} and compare them to a standard trajectory controller. 
We run experiments to show the effectiveness of the proposed controller in terms of object displacement minimization and external force compensation.
In a set of ablation studies we show the impact of each controller component on its behavior.

\section{Related Work}

In a review of human grasping, Johansson and Flanagan~\cite{johansson2009coding} highlight the importance of tactile sensations.
They divide a grasping task into different phases, where most of the phase transitions are determined by different tactile events.
When anticipating object contact, during the reach motion, humans perform position control.
Once object contact is detected, we switch to the load phase where the grasp force is increased until a goal force is reached.
During subsequent phases, such as lifting, holding, and replacing, we try to maintain that goal force.
In the unload phase, we gradually decrease the contact force until we release the fingers.
These findings are supported by experimental studies such as~\cite{Sundaram2019}.

The human grasping approach presented in~\cite{johansson2009coding} has also been adopted by many works in robotics.
Romano et. al.~\cite{romano2011human} closely follow the phase distinctions from \cite{johansson2009coding} while modelling their force controller for the PR2 robot.
They concentrate on goal force estimation and slippage prevention using a small sensor array in the 1-DoF PR2 gripper.
One of their assumptions is the equidistant placement of each finger around the object, thus implicitly assuming force closure on first contact.
In an earlier work, Hsiao et. al. \cite{Hsiao2010} also proposed a force controller for the PR2, but without the assumption of equidistant finger-object placement.
If one finger acquires object contact first, they move the arm into the direction of that finger in order to avoid object movements and guarantee force closure.
Their work was integrated and evaluated in a grasping pipeline for the PR2 in \cite{Ciocarlie2014}.
Since our work focuses on controlling grippers with 2-DoF, we need to impose additional constraints on the controller to avoid object drift.
This allows us to be compliant with external forces, which is not possible with a 1-DoF gripper.

Many works exist that consider grasp force control in fixed-arm setups with 2-DoF \eefs.
The authors in \cite{Patel2018} designed a proximity, contact, and force sensor using infrared sensors behind an elastomer.
They leverage the proximity measurements to center the gripper around the object and ensure a safe grasp this way.
Similar to \cite{romano2011human}, the authors process their sensed forces to match human tactile perception channels, as described in~\cite{johansson2009coding}.

The studies \cite{Su2015} and \cite{montano2018} use two fingers of a multi-fingered hand to perform force control.
In \cite{Su2015}, the authors classify slippage using BioTac sensors. 
Once object slip is detected, they increase the desired grasp force, thus stabilizing the grasp.
\cite{montano2018} concentrates on optimizing grasp stability during the holding phase by updating fingertip positions based on force closure constraints.
In contrast to our work, these studies do not take the effect of external object forces on the controllers into account.

A number of authors also aim for multi-fingered hands, which poses additional challenges.
\cite{tahara2010dynamic} propose a control scheme for a three-fingered hand by defining and controlling a virtual object frame that is based on finger contacts.
In \cite{li2012towards}, the authors propose a combined position and force controller to ensure stable grasps for five-fingered hands that is based on \cite{tahara2010dynamic}.
While both studies point out the need to consider external forces such as gravity, the evaluations were conducted in simulation without taking external forces into account.

More recent studies have evaluated their methods for multi-fingered hands also on real hardware.
\cite{Hang2016} propose a grasp stabilization approach for the Allegro hand, that controls a virtual object frame through force adaptation and finger gaiting.
In \cite{Deng2020}, Neural Networks and Gaussian Mixture  Models are used to detect contacts and estimate forces on the Shadow Hand.
The results are subsequently used by a grasp-stabilizing controller that adjusts finger positions accordingly.
In both works, the resilience to external disturbances is evaluated, either by a human, shaking the arm (\cite{Deng2020}) or by the arm executing a shaking motion (\cite{Hang2016}).
However, these experiments do not investigate the effect of constant and maintained forces on the fingers, such as gravity or humans pushing or pulling on the fingers.
Our controller is modelled to explicitly handle both types of external forces, peaky disturbances and slow continuous drift.

A growing subfield in the tactile manipulation community is that of reinforcement learning.
Some approaches aim to learn stable grasping policies \cite{Merzic2019}, while others focus on increasing grasp success using tactile feedback \cite{Wu2019}.
One challenge of these approaches is the sim-to-real transfer \cite{church2021tactile}, since modelling tactile sensor responses is non-trivial in simulation~\cite{akkaya2019solving}.
External forces and their physical effects are also difficult to model.
Hence, most machine learning related works do not take external forces into account.
Our controller could be used as a component of a more complex machine learning model to address these issues, provided that the output of the model are desired forces.
\section{Grasp Controller}
\begin{figure}
    \centering
    \begin{tikzpicture}[->,>=stealth',auto,node distance=3.2cm,align=center]
        \tikzstyle{every state}=[draw,rectangle,rounded corners=10pt,
                                 minimum size=1.5cm]
        \node[state] (A)               {I\\{\small closing}};
        \node[state] (B) [right of=A]  {II\\{\small establish}\\{\small contact}};
        \node[state] (C) [right of=B]  {III\\{\small holding}};
        \path (A) edge node {$f_i > \ft$}   (B)
              (B) edge node {force\\closure} (C);
    \end{tikzpicture}
    \caption{State diagram illustrating the 3 phases of grasping.}
    \label{fig:states}
\end{figure}
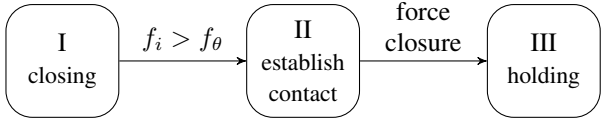

We propose a holistic control approach for grasping, covering all phases from finger closing over establishing contacts to holding (cf.~Fig.~\ref{fig:states}).
Consequently, the controller distinguishes these three phases and applies a different control strategy in each of them, similar to human grasping as described in~\cite{johansson2009coding}.
During phase I, when the fingers are approaching the object, the controller performs open-loop trajectory control by following a pre-planned joint trajectory targeting a hand posture that would eventually penetrate the object.
Once a tactile sensor reliably detects contact, i.e.\ perceives a contact force $f(t)$ just above the noise threshold $\ft$, the corresponding finger transitions to phase II.
As a result, all joints along the kinematic chain from the sensor link up to the hand base stop moving but just maintain object contact. 
This avoids undesired object movements during grasping in situations where the object is not perfectly centered between the fingers.

Each time a new contact is established, the resulting grasp is evaluated for force-closure~\cite{murray1994mathematical} as described in sec.~\ref{sec:force-closure}. 
If the grasp is considered force-closure, phase III is entered, which finally performs grasp force control aiming to establish and maintain a desired grasp force as outlined in sec.~\ref{sec:force-control}. 
We assume desired goal forces to be provided by a higher-level cognitive system component based on common-sense object knowledge, which is analogous to how humans decide which grasping force to use~\cite{johansson2009coding}.
Alternatively, the grasp forces could be reactively adapted due to recognized incipient slippage as suggested e.g.\ in \cite{romano2011human, Hang2016, haschke2015grasping,walck2017fatigue}.

\subsection{Evaluating Force-Closure}\label{sec:force-closure}
Force-closure is a formally defined property of a grasp configuration stating that the grasp can balance any (bounded) external disturbance wrench with contact forces $\vec f_i$ that satisfy the friction cone constraints, i.e. don't induce local slippage.
To determine the net wrench applied onto an object through contact points $\vec p_i$, independent contact forces $\vec f_i$ are mapped onto 6-dimensional contact wrenches $\vec F_{C_i} = (\vec f_i, \tau_i)$ expressed in local contact frames $C_i$, which are then transformed into the common object frame $O$ using the adjoint matrix \adj, and finally summed up to yield the net wrench $\vec F_O$ expressed in frame $O$ \cite{murray1994mathematical}:
\begin{align}
    \vec F_O &= \sum_{i=1}^n \adj \cdot \vec F_{C_i} = \sum_{i=1}^n \adj B_i \cdot \vec f_i 
            \equiv G \cdot \vec f \\\notag
    B^\T &= \left[\begin{smallmatrix}
        1 & 0 & 0  & 0 & 0 & 0 \\
        0 & 1 & 0  & 0 & 0 & 0 \\
        0 & 0 & 1  & 0 & 0 & 0 \\
        0 & 0 & 0  & 0 & 0 & 1 \\
    \end{smallmatrix}\right]
    \quad \adj = \begin{pmatrix}
        R_\oc \\ 
        \hat{\vec p}_\oc \cdot R_\oc && R_\oc
    \end{pmatrix}
\end{align}
Here, $\vec f = [\vec f_1^\T, \dotsc, \vec f_n^\T]\T$ denotes the vector of concatenated contact forces $\vec f_i$, $G$ denotes the grasp matrix, and $B \equiv B_i$ is the wrench basis modelling soft-finger contacts \cite{murray1994mathematical}.
The latter is a point-contact model allowing for \emph{force} transmission along the contact normal ($f_{i,z}$) and within the tangential plane ($f_{i,x}, f_{i,y}$) as well as for \emph{torque} transmission along the normal axis of the contact ($f_{i,\tau}$). These independent force components $\vec f_i \in \mathbb{R}^4$ need to satisfy friction cone constraints $\FC_i$ derived from Coulomb friction to avoid slippage:
\begin{equation*}
    0 \leq \sqrt{f_{i,x}^2 + f_{i,y}^2} \leq \mu_i f_{i,z}
    \quad\text{and}\quad |f_{i,\tau}| \leq \mu_{i,\tau} f_{i,z} \,.
\end{equation*}
A grasp defined by its grasp matrix $G$ and the collection of all friction cone constraints $\FC = \FC_1 \times \dotsc \times \FC_n$ is force-closure, iff $G$ is surjective and there exist strict internal forces, i.e. forces $\vec f_N$ that strictly satisfy the friction cone constraints and that don't have a net effect onto the object ($G \cdot \vec f_N = 0$). Both conditions can be easily verified given the geometry of the grasp, i.e. its contact locations $\vec p_i \equiv \vec p_{oc_i}$, contact normals $\vec n_i$ (determining $R_{oc_i}$), and a conservative estimation of the friction coefficients $\mu \equiv \mu_i$.

\subsection{Grasp Force Control}\label{sec:force-control}
We assume that the tactile sensors only provide contact forces along the normal direction at the contact point.
This rather weak assumption allows applying the grasping force controller to a large variety of tactile sensors \cite{kappassov2015tactile}, but limits the capabilities of the method.
Traditional grasp force controllers like~\cite{tahara2010dynamic} strive for a globally optimal contact force distribution ensuring grasp stability,
i.e. ensuring that (i) all contact forces stay within friction cone bounds,
(ii) applied forces exactly balance external forces (e.g. gravity), and (iii) local contact forces are minimized \cite{murray1994mathematical}.
However, this approach is only meaningful if the full 3D contact force is observable and thus controllable.
Here, we assume that we cannot measure shear forces nor determine the friction coefficients required to evaluate friction cone constraints.

Consequently, we are not aiming at realizing a net wrench of zero applied onto the object as proposed in traditional grasp force controllers.
Rather, we restrict control to the observable wrench component only, which in our case are normal forces along the gripper's closing direction $\vec n$.
Hence, instead of aiming for $G \cdot \vec f + \vec F^\ext \!=\! 0$, we project onto the normal direction and yield:
\begin{equation}
    (f_1 \vec n - f_2 \vec n + \vec f^\ext) \cdot \vec n = f_1 - f_2 + \fext \overset{!}{=} 0 \,,
    \label{eq:ext-force}
\end{equation}
where $f_i$ denote scalar normal force components measured by the two fingertip sensors and $\fext$ is the scalar external force component along the normal direction.

To maintain a desired grasp force $f_\g$, we consider the sum of both contact forces, which essentially resembles the overall magnitude of \emph{internal} forces applied to the object and thus should become equal to $f_\g$: 
\begin{equation}
    \fint \equiv f_1 + f_2 \overset{!}{=} \fgoal
    \label{eq:int-force}
\end{equation}
By decoupling control of external and internal forces in this fashion, we decouple object motion from object deformation. In contrast, the naive approach controlling both contact forces $f_1$ and $f_2$ independently, would face a common practical issue: in real-world settings, force measurements are noisy and not perfectly calibrated.
This means, that even given balanced antipodal force \emph{measurements}, the \emph{actual} physical forces might differ, thus resulting in a net force applied to the object. 
This imprecision needs to be considered in order to avoid object drift when no
external forces are applied.
Note that simple 1-DoF grippers, like in \cite{romano2011human}, do not face this issue: their fingers move in sync and the object will always be centered in the gripper.

When the measured imbalance of forces is larger, we expect the cause to be an external force being exerted on the gripper.
One type of external force are those resulting from physical interactions with the environment, e.g. pushing or pulling on the fingers or collisions with the environment.
Since the controller’s main objective is to maintain and not exceed $\fgoal$, it needs to be compliant with these forces.
Another external force influencing the sensors measurements is gravity. As we don't want the fingers to drift due to gravity, we need to explicitly consider gravity in eq.~\ref{eq:ext-force}, resulting in:
\begin{equation}
    f_1 - f_2 + m\,\vec g \cdot \vec n = -\fext \overset{!}{=} 0
    \label{eq:ext-gravity}
\end{equation}
where $m$ is the object's (and sensor's) mass in kilograms, $\vec g = 9.81 \frac{m}{s^2} \, \vec z$ is the acceleration vector due to gravity, and $\fext$ is the remaining external force. This formulation emphasizes the fact that we cannot explicitly counteract external shear forces, but only normal forces.

In this work, we further assume that the end-effector is position-controlled. Consequently, we transform force errors into Cartesian position errors according to Hooke's law:
\begin{equation}
    \label{eq:xin}
    \Dx(t) = \frac{1}{\ks} \Delta f(t) \,.
\end{equation}
Here, \ks is the stiffness constant, i.e. in case of eq.~\ref{eq:int-force} the object's stiffness and in case of eq.~\ref{eq:ext-gravity} the desired compliance when giving in to external forces.
Estimating the object's stiffness accurately is difficult in practice. Thus we utilize PI control to account for uncertainties in estimating $\ks^\inner$:
\begin{equation}
    \label{eq:uin}
    u^\inner(t) = \frac{\kp^\inner}{\ks^\inner} \cdot \Delta f^\inner(t) + \frac{\ki^\inner}{\ks^\inner} \int_\text{III} \!\!\Delta f^\inner(\tau) \, d\tau \,,
\end{equation}
where $\kp^\inner$ and $\ki^\inner$ denote the coefficients of the proportional and integral terms.
The integral is calculated over all time steps since the transition into controller phase III.
Thus, for non-zero force errors $\Delta f^\inner = f_1 + f_2 - f^\g$ the integral term will increase in magnitude and thus force convergence of the error towards zero.

To react to external forces, expressed via eq.~\ref{eq:ext-gravity}, we employ a simple P-controller:
\begin{equation}
    u^\ext(t) = \frac{\kp^\ext}{k^\ext}\cdot \fext \qquad \text{if } |\fext| > f_\phi
    \label{eq:ext-compliance}
\end{equation}
where $f_\phi$ is a small threshold to filter out sensor noise, similar to $\ft$ for the force readings.
To avoid object drift caused by sensor noise or poor calibration, this compliant position controller is only applied if $|\fext|$ is larger than $f_\phi$.
Inside this deadband, we perform position control to maintain the object position $O(T)$ observed at the last time $T$ when true external forces were measured:
\begin{equation}
    u^\ext(t) = \Delta O(t) \equiv O(T) - O(t) \qquad \text{if } |\fext| \leq f_\phi
    \label{eq:ext-position}
\end{equation}

In practice, we set $\fte = \ft$, however it is important for the experiments to distinguish between \fte and \ft.
This controller component can be thought of as a constraint on a virtual object frame, an idea that has been pursued before \cite{tahara2010dynamic, Hang2016}.
Our method differs from previous approaches since we condition this object pose constraint on values of \fext while taking gravity into account.

Finally, joint position deltas are computed via inverse velocity kinematics. In our case, where increasing joint positions correspond to an opening of the fingers, we yield:
\begin{align}
    \Delta q_1 &= {\textstyle\frac{1}{2}}(-u^\ext + u^\inner) \\
    \Delta q_2 &= {\textstyle\frac{1}{2}}(+u^\ext + u^\inner) \,,
\end{align}
which distributes control signals evenly between the two fingers. The commanded joint positions are determined by simple integration:
\begin{equation}
    \vec q^\cmd(t) = \vec q(t) + \Dq(t) \,.
\end{equation}

\subsection{Implementation}
We implemented the grasp controller using the robot operating system (ROS) \cite{quigley2009ros}, the most commonly used robotics middleware nowadays.
The default controller library ros\_control \cite{rosctrl} offers a JointTrajectoryController (JTC), which performs free position control for a set of joints given a trajectory.
Our controller inherits from the JTC class, keeping all its interfaces.
Hence, a grasp can be initiated by sending a JointTrajectoryActionGoal in the same fashion a JTC task would have been triggered.
Therefore, our controller serves as a drop-in replacement for existing grasping pipelines.
Parameters specific to force control can be comfortably adjusted via the dynamic reconfigure interface.
The code is open-source and can be found on GitHub\footnote{\url{https://github.com/llach/ta11_sensor_tools}}.

The controller offers two different modes for phase III: an open-ended force control mode and a mode finishing control upon reaching the desired force. 
For the former one, the controller attempts to hold the desired force forever.
During this mode, the desired force can be adapted, e.g.\ reactively in response to incipient slip detection or deliberately prior to planned interactions with the environment.
The latter mode signals action success once $f^\g$ is reached.

\section{Experiments}

We evaluated the control approach on the TIAGo robot, a service robot developed to serve in household settings.
It has a 7 DoF arm with end-effectors that can be easily exchanged or modified, making it an extensible and adaptable manipulation platform.
Since our work is designed for two DoF \eefs, we chose the \pjg for our experiments.
With the gripper, TIAGo has a maximum payload of $2.1kg$ if the arm is fully extended.
By default, TIAGo's gripper has two plastic fingers that are \emph{not} sensorized.
For our experiments, we swapped the fingers for load cell sensors (TA11 from Gicam) with 3D-printed, rigid plastic fingertips as shown in Fig.~\ref{fig:grasp}.
Using the sensors, we can read contact forces normal to the gripper's closing direction from each finger.

The raw sensor values are read using a LabJack U6 and then published in a bias-corrected fashion to a ROS topic.
To this end, the ROS node estimates the sensor biases $b_i$ by averaging over $1000$ measurements in unloaded state on each startup.
Additionally, force measurements are also converted into Newton.
The conversion factors $\gamma_i$ were determined by aligning the sensors parallel to the ground, recording de-biased sensor responses for three objects with different weights, calculating their individual conversion factors and then taking the mean of the three factors.
This way, we determined $\gamma_1 = 11.02$ and  $\gamma_2 = 11.03$.
Calibrated forces are then calculated by:
$$
f^{\text{calibrated}}_i = \gamma_i \cdot (f^{\text{raw}}_i - b_i)
$$
Furthermore, we determine the noise threshold $\ft=0.2N$ by observing the maximum value during the bias correction process and approximately doubling it.
We choose it to be relatively conservative in order to avoid false-positive contact detections.
Choosing a well-suited value is crucial for the controller's performance.
A too tight threshold can lead to premature stopping of the finger-closing phase due to noise, while a too conservative threshold degrades the controller's sensitivity.

To assess the sensor's first-touch sensitivity, we aligned them parallel to the ground and placed small weights (different Euro coins) on their tips
until the measured sensor value was continuously above the threshold \ft for a period of 30 seconds.
In this fashion, we determined first-touch sensitivities of $0.35N$ and $0.38N$ for the left and right sensors,  respectively.

Prior to the experiments, we determined the controller parameters in preliminary grasping trials.
The PI controller parameters as well as the object stiffnesses were chosen such that objects were grasped without overshooting the goal force.
This can lead to a slightly slower convergence to the goal force but ensures that the object is not damaged.
Consequently, we chose the PI coefficients $\kp^\inner = 1.9$ and $\ki^\inner = 9.0$.
We considered three different objects: a styrofoam cylinder of 2$g$, a tape roll of 49$g$ and a wooden cuboid of 144$g$.
For all experiments, $k$ was set to 1000.
The estimation process for $k$ has proven to be rather quick, requiring only about five trials per object.
We found that tuning it for soft objects is usually sufficient as the controller's integral part will still lead to stable grasps of stiffer objects.

\subsection{Object Displacement}

In real-world scenarios, the \eef is rarely centered perfectly around the object when executing a grasp, resulting in object displacements if initial object contact cannot be detected reliably.
With experiment \one, we analyze our controller's behavior in these situations and compare it to standard position control.

\begin{figure}
    \includegraphics[width=0.49\linewidth, trim=0 0 0 200, clip]{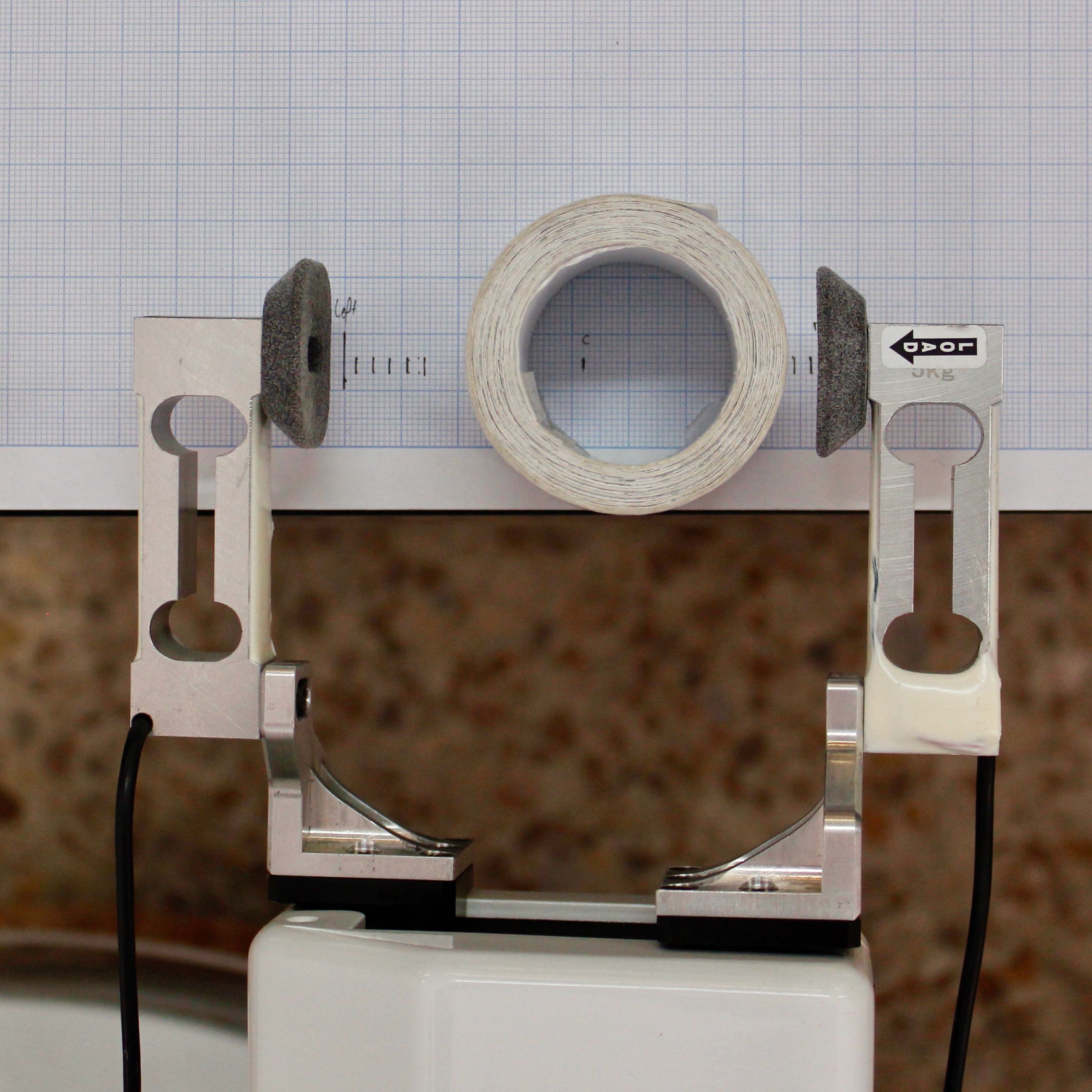}\hfill
    \includegraphics[width=0.49\linewidth, trim=0 0 0 200, clip]{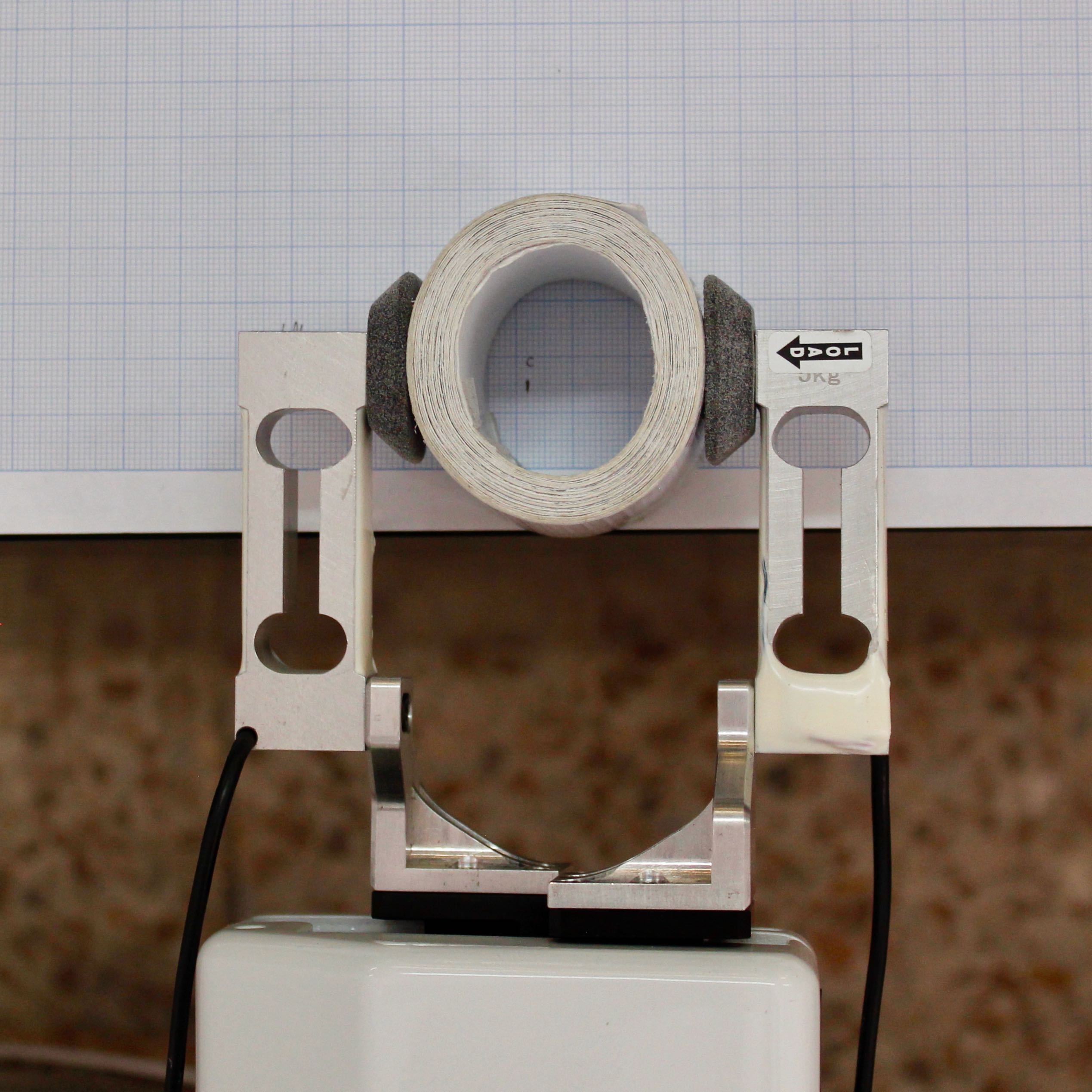}
    \caption{Setup for experiment \one: objects are placed with an offset relative to one finger (left) and then grasped (right). After the grasp, the object displacement is measured.}
    \label{fig:exp1}
\end{figure}

For the experiment, we position the open gripper over a table and place an object between its fingers.
When placing the object, we do not center it perfectly, but rather place it closer to one of the fingers.
The gripper is then closed, either with our force controller or with position control.
Finally, the distance the object was moved during the gripper closing is recorded.
This process is repeated for the three aforementioned objects at five different object positions three times.
The object offsets $o_i$ of $2mm$, $5mm$, $8mm$, $11mm$ \text{and} $14mm$ were chosen such that all three objects would fit into the gripper at each offset.
In total, $5 \cdot 3 \cdot 3 = 45$ grasping trials were performed for each controller.
The experimental setup is shown in Fig.~\ref{fig:exp1}.

\subsection{Grasp-force controller -- ablation studies}
With experiment \two, we evaluate the utility of the individual controller components introduced in sec.~\ref{sec:force-control}. 
In particular, we perform ablation studies where we evaluate grasping behavior while one of the controller components is switched off.
In this fashion, we evaluate the utility of the decoupling of internal and external force control, the conditional external force compensation, and the gravity compensation.

First, we study two different grasping scenarios with all components enabled.
In the first scenario, external forces are exerted on the gripper's fingers by a human while holding an object.
Here, the controller is expected to be compliant with the forces, but not to drift once they are removed.
In the second scenario, the gripper is being moved such that the influence of gravity on the held object is changing.
After evaluating the controller's behavior with all components enabled, we re-run these scenarios, but with individual components switched off.
As a result, we can qualitatively assess the utility of a component.

\section{Experimental Results}


\begin{table}
    \centering
    \begin{minipage}[c]{\linewidth}
        \centering
        \caption{Comparison of object displacements in $mm$ for standard position control (JTC) and our force controller. Each object was grasped $3$ times from $5$ different finger offsets to simulate imperfect gripper-object alignment. Displacements over all offsets were averaged.}
        \begin{tabular}{ccccc} \toprule
            \textbf{Object} & \textbf{Weight} & \textbf{JTC} & \textbf{Our Controller}\\\midrule
            Styrofoam & $2g$ & $19.0 \pm 4.2 $ & $17.0 \pm 4.2 $ \\ 
            Tape roll & $44g$ & $21.7 \pm 4.5 $ & $\textbf{3.5} \pm 0.5$  \\ 
            Wooden cuboid & $144g$ & $9.0 \pm 4.2 $ & $\textbf{1.0} \pm 0.3 $ \\ 
            \bottomrule
        \end{tabular}
        \label{tab:displacement}
    \end{minipage}
    \vspace{-0.5cm}
\end{table}

\subsection{Object Displacement}
For experiment \one, we used three objects with different weights.
We measured the object displacements by taking the difference of finger positions before and after grasping minus the object offset: $q^{\text{start}}-q^{\text{end}}-o_i$.
Tab.~\ref{tab:displacement} compares the displacements of the objects for a series of 15 grasps for both, force and position control.
For the styrofoam cylinder, the displacement for both controllers is very similar: 
this object was too lightweight to trigger the first-touch detection on the force sensors.
For the other two objects, the displacement was significantly reduced by using our force controller.
The experiments show that for sufficiently heavy objects, our controller minimizes object displacement to less than $1mm$.
Furthermore, the results show that the object displacement is equally minimized across different object offsets, as indicated by the low standard deviation measured for the tape roll and the cuboid.


\begin{figure}
    \centering
    \includegraphics[width=0.49\textwidth]{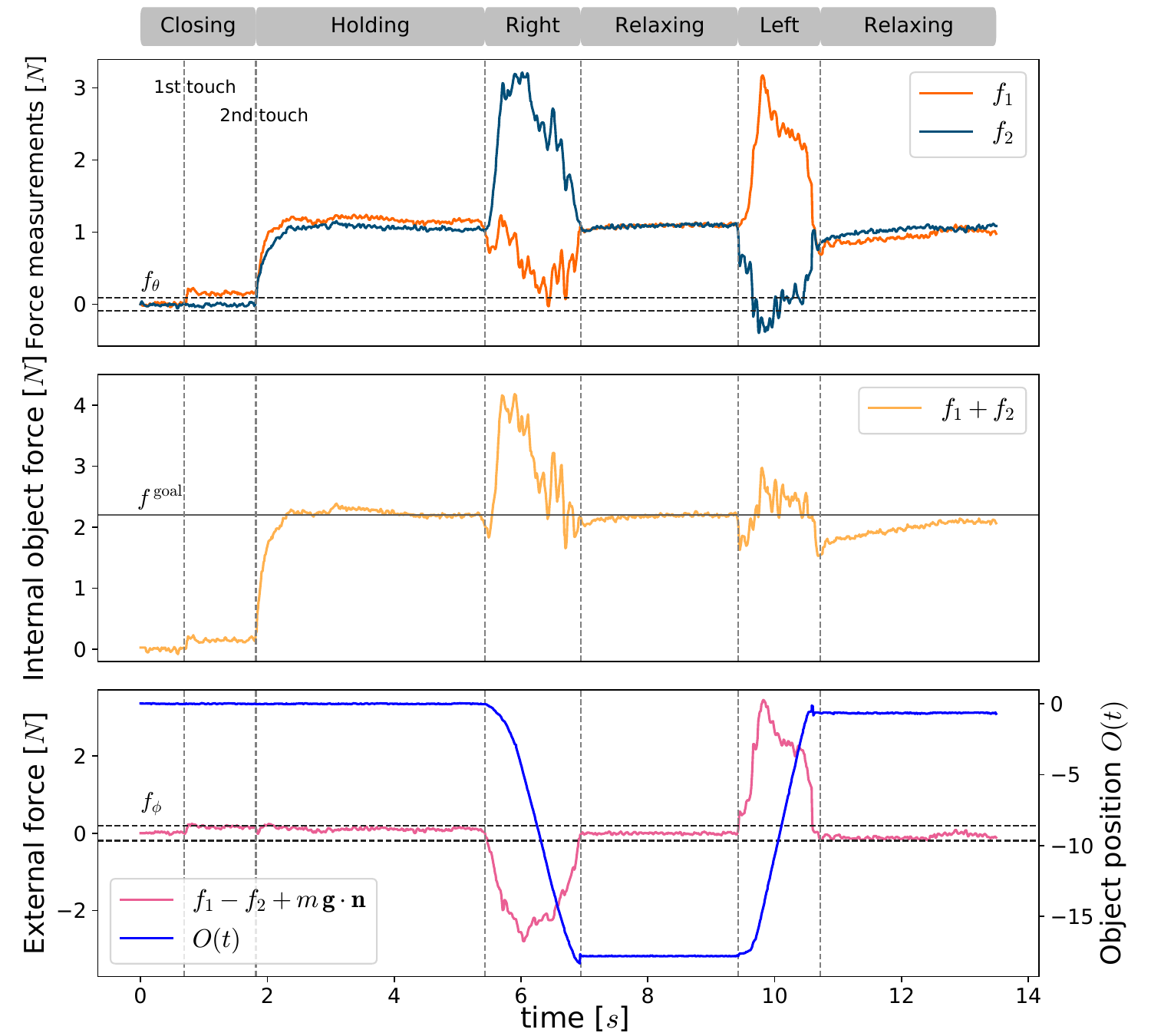}
    \caption{Controller behavior during object grasping (tape roll) and application of external forces, first from the right and then from the left.}
    \label{fig:external_forces}
\end{figure}

\begin{figure}
    \centering
    \includegraphics[width=0.48\textwidth]{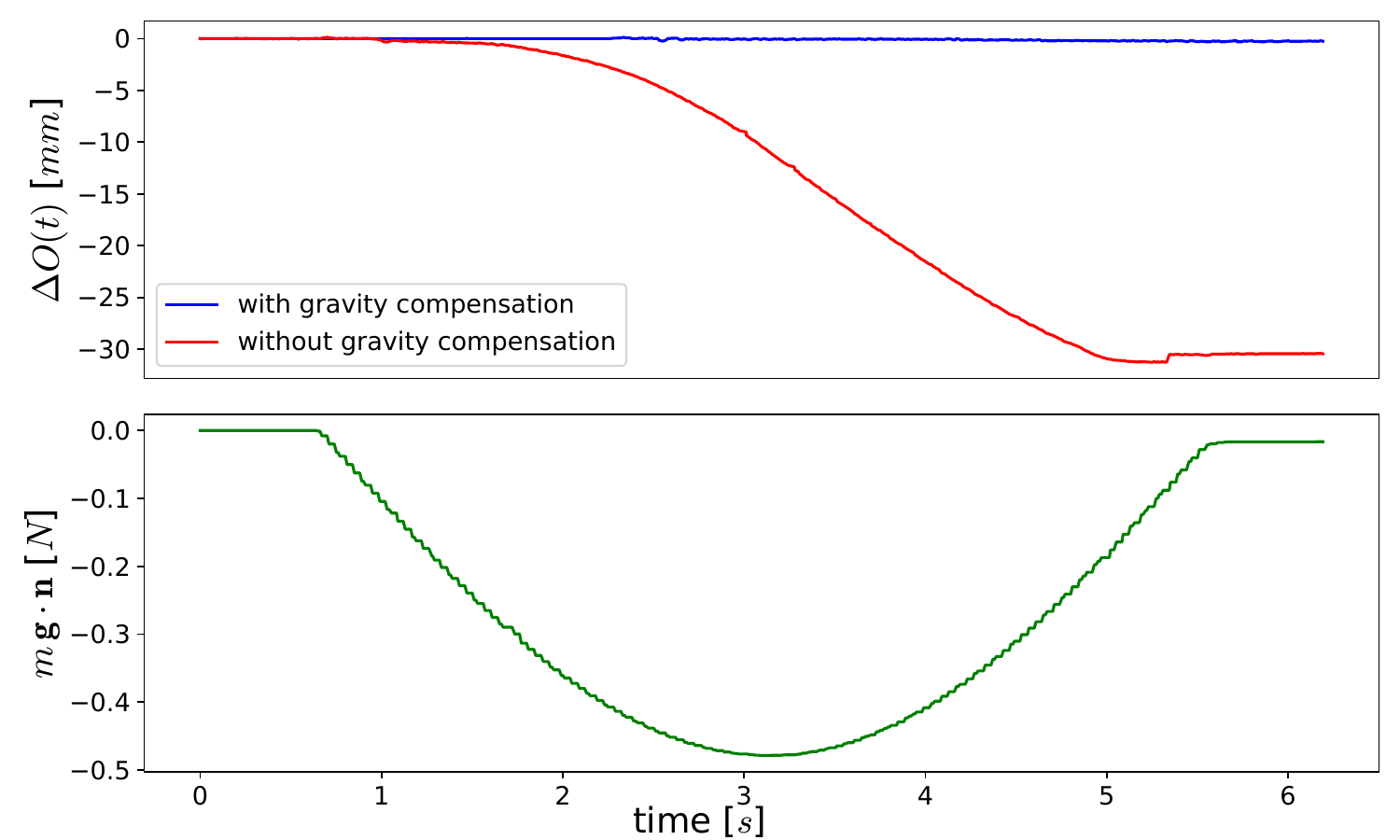}
    \caption{Object drift during gripper rotation while holding the tape roll, with and without gravity compensation.}
    \label{fig:gravity}
    \vspace{-0.5cm}
\end{figure}

The results from experiment \one are twofold: it shows, that using our force controller, we can minimize undesired side effects of object manipulation during the grasping stage.
By dividing a grasp into different phases as humans do, object displacements were minimized and force closure was guaranteed.
Furthermore, it shows that the performance of any force controller highly depends on the sensors' capability to detect first touch.
That capability is in turn dependent on the sensor sensitivity as well as its calibration.
As hardware continuously improves, we expect to see more sensitive tactile sensors in the future.
A higher sensitivity would decrease object displacements even more and enable TIAGo to handle lighter objects.

\subsection{Grasp-force controller -- ablation studies}
For the first scenario of experiment \two, we performed a grasp using our force controller with the gripper being parallel to the floor and all components enabled to establish a baseline.
After initial grasping, we hold the object without any external forces to verify that no object/gripper motion is occurring.
Subsequently, we applied external forces to one finger.
In response, the controller compliantly gave in as expected and started moving away from the force.
Once the external force was taken away, the finger movement stopped promptly, and the object's position was stable again.
The controller reacted the same when applying forces on the other finger.
Fig.~\ref{fig:external_forces} shows the internal and external object forces as well as the object position that were measured for this scenario.
The different phases and events, such as initial contacts, are annotated as well.

For the first ablation study, we disabled compliance to external forces (setting $\fte = \infty$ in eq.~\ref{eq:ext-compliance}).
As a result, the gripper could not give in to external forces.
This caused the total force exerted on the object to surpass \fgoal when external forces were applied.
Obviously, this is not a desired behavior as fragile objects might get damaged.

In the second ablation study, we set $\fte = 0$, disabling object position control within the deadband.
While the controller was able to give in to external forces, the gripper drifted away even after the force was removed.
This is likely caused by tiny differences in force measurements between both sensors.
Enabling the deadband \fte results in the desired behavior of stopping the motion promptly after taking away the external force.

In the second scenario, we rotated the wrist by approximately 180 degrees after grasping an object.
Consequently, the sensor's normal vector changed with respect to the gravity vector.  
During the first 90 degrees, the sensor normals became aligned to gravity, hence maximizing its influence.
For the remaining rotation, the influence of gravity decreased again and became nearly zero.

In the corresponding ablation study, we evaluated the controller behavior when not explicitly compensating for gravity.
Thus, \fext was \emph{not} corrected using $m \, \vec g \cdot \vec n$ and the controller confused gravity for an external force.
This caused the gripper's finger to drift towards the ground when the gripper was rotating.
Fig.~\ref{fig:gravity} compares the object displacement with and without gravity compensation while performing the described movement.
Without gravity compensation, the object drifted as much as 30 $mm$, while the object drift was negligible ($< 1mm$) with gravity compensation enabled. 

In conclusion, these ablation studies have demonstrated the importance of each controller component.
If one was disabled, either undesired object drift was occurring or \fgoal was surpassed.
We also compare the controller's behavior with and without the individual components side-by-side in an accompanying video, which illustrates these comparisons in an accessible manner.
\section{Conclusion And Future Work}
In this work, we have presented a holistic grasp controller that combines free position control during finger closing with force control during object holding.
We proposed a controller that minimizes undesired object displacement while holding a pre-determined desired force.
At the same time, the controller is compliant with external forces in order not to surpass the goal force.
Furthermore, gravity as an external force is explicitly accounted for to avoid undesired object drift.

We have validated our claims in two experiments:
In experiment \one, we have shown our controller's ability to minimize object displacement during grasping and execute force closure grasps even under object pose uncertainty.
The ablation studies of experiment \two have illustrated how each component is contributing to the final behavior of the controller. 
Each component was shown to be crucial to either minimize undesired object movements or to be compliant with external forces only when they are actively applied to the fingers.

Although demonstrated for a simple \pjg only, we think the approach will easily generalize to multi-fingered hands as well.
To this end, the general concept of decoupling control of external and internal forces needs to be generalized,
 considering the full 6-dimensional wrench space and the nullspace of the grasp matrix respectively.
We plan to implement this generalized version of the controller in our future work.

\renewcommand*{\UrlFont}{\rmfamily}
\printbibliography
   
\addtolength{\textheight}{-12cm}   

\end{document}